\documentclass{article} 
\usepackage{iclr2027_conference,times}

\usepackage{amsmath,amsfonts,bm}

\def\eqref#1{equation~\ref{#1}}

\def\1{\bm{1}}

\DeclareMathAlphabet{\mathsfit}{\encodingdefault}{\sfdefault}{m}{sl}
\SetMathAlphabet{\mathsfit}{bold}{\encodingdefault}{\sfdefault}{bx}{n}

\usepackage{hyperref}
\usepackage{url}
\usepackage{amssymb}
\usepackage[table]{xcolor}
\usepackage{pifont}
\usepackage{multirow}
\usepackage{graphicx}

\title{Learning to Restore More: Continual Capability Expansion for Pretrained Image Restoration Models}

\author{Hu Gao \& Lizhuang Ma \\
Department of Computer Science \\
Shanghai Jiao Tong University \\
Shanghai, China \\
\texttt{\{gao\_h, lzma\}@sjtu.edu.cn} \\
\And
Yulong Chen \\
 Department of Architecture and Design \\
Harbin Institute of Technology \\
Heilongjiang, China\\
\texttt{\{llong\_c\}@hit.edu.cn} \\
}

\iclrfinalcopy 
\begin{document}

\maketitle

\begin{abstract}
Image restoration models are typically trained with a fixed set of capabilities. 
When new restoration requirements emerge, existing solutions usually train additional models or jointly retrain the original model with both new and historical data. 
Instead of designing another restoration backbone, we investigate how a trained restorer can continually acquire new capabilities without forgetting those learned previously. 
We propose \textbf{RestoreMore}, a continual capability-expansion framework that preserves the pretrained restoration model as a frozen capability anchor and learns residual expansion modules for newly arriving degradations. 
RestoreMore introduces a capability-oriented bi-level routing mechanism at multiple feature stages. 
The first routing level identifies restoration capabilities relevant to the current input, while the second selects and combines a sparse set of complementary degradation experts. 
This design enables newly introduced tasks to selectively reuse historical
restoration knowledge and progressively enriches the expert bank available
for subsequent restoration tasks.
Extensive experiments on a wide range of restoration benchmarks demonstrate that RestoreMore consistently acquires new restoration abilities while preserving and improving previously learned capabilities.

\end{abstract}

\section{Introduction}
\label{sec:introduction}

Image restoration aims to recover high-quality images from observations corrupted by noise, rain, and other adverse factors. 
Most existing methods, however, are developed under a fixed-capability assumption. A degradation-specific model~\cite{LSSRgao2024learning, efderainguo2025efficientderain+,202611333962,11342300} is optimized for a predefined task and usually performs poorly when the degradation changes. All-in-one image restoration alleviates the need to maintain separate networks by learning multiple restoration tasks within a unified model~\cite{CAPTNet10526271,Allrestor11367271,baryir11417902,gao2026learning}. Nevertheless, the capability set of an all-in-one model is still determined by the degradation types available during its original training.

\begin{figure*}
    \centering
    \includegraphics[width=\linewidth]{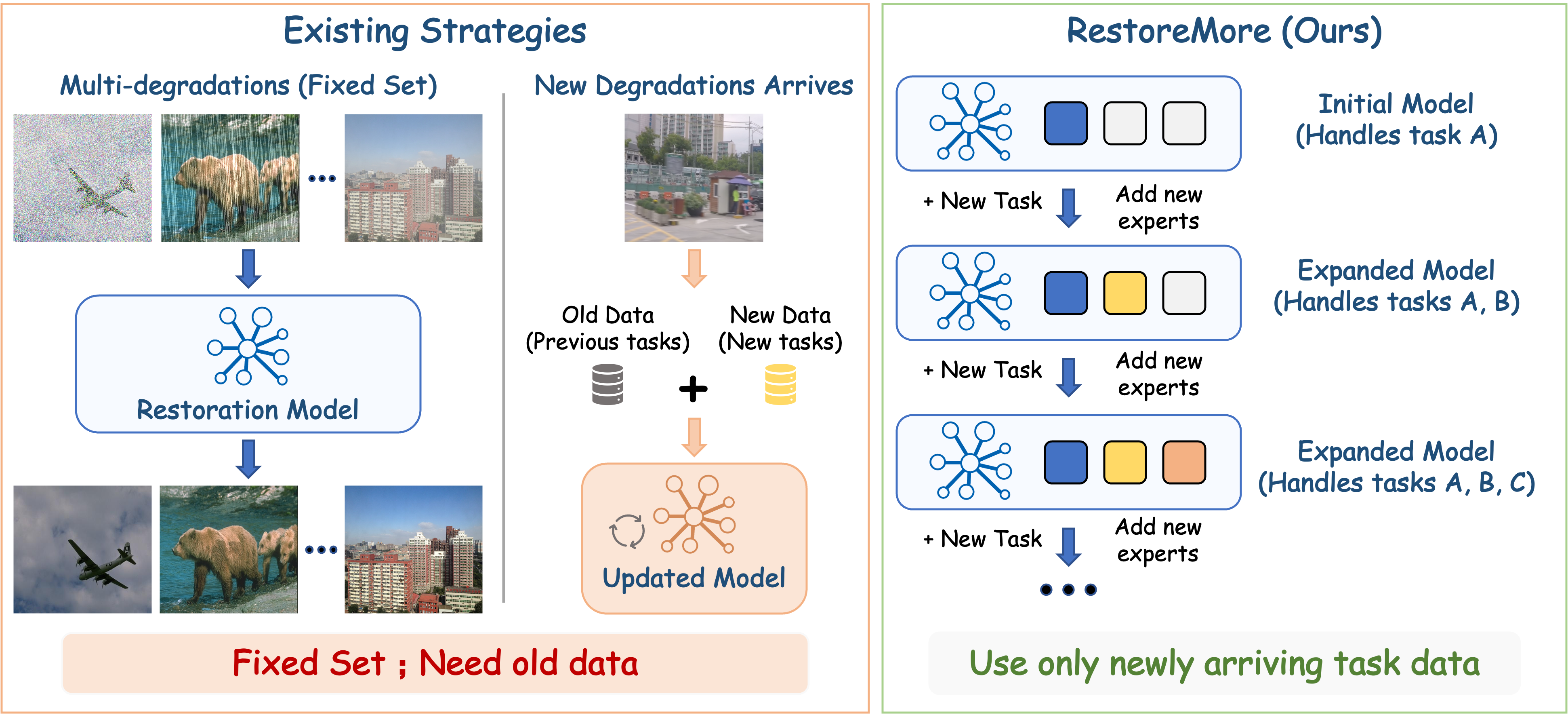}
 \caption{
Comparison of existing restoration strategies and RestoreMore.
Existing methods are limited to a fixed degradation set and typically require joint retraining with historical and new data when new degradations arrive.
In contrast, RestoreMore continually expands a pretrained restorer by adding new experts using only newly arriving task data.
}
    \label{fig:intro_comparison}
\end{figure*}

This closed-set assumption limits the long-term utility of trained restoration models. 
A denoising model may subsequently be required to remove rain or haze, while an all-in-one model originally trained for noise, rain, and blur may later encounter snow or previously unseen imaging artifacts. 
A common solution is to train an additional task-specific model or jointly retrain the original model using the expanded collection of new and historical datasets. 
The former continually increases storage and deployment costs, whereas the latter requires access to all previous data and repeatedly optimizes knowledge that has already been learned. Fig.~\ref{fig:intro_comparison} illustrates the difference between these existing strategies and our capability-expansion perspective.

Several studies have explored continual learning for image restoration~\cite{cheng2024continual, lu2025continuous, gong2025minimalistic}, demonstrating the feasibility of incrementally extending restoration capabilities.  However, existing approaches are generally built on dedicated incremental architectures or restricted restoration task families, with their primary objective being the mitigation of catastrophic forgetting. They do not explicitly address how to preserve the restoration pathway of
an arbitrary pretrained model or how to selectively reuse complementary
knowledge accumulated from different degradation types and feature stages.

In this paper, we study image restoration from a continual capability-expansion perspective. 
Rather than constructing another restoration backbone for a predetermined task collection, we ask a different question: given an already trained image restoration model, how can its restoration capability be continually expanded to previously unsupported degradations? 
After learning a new degradation, the expanded model should acquire the corresponding restoration ability while retaining the capabilities learned before expansion. 
Meanwhile, the accumulated expert bank should enable knowledge reuse across
related degradation types rather than treating each newly arriving task in isolation.

To achieve this objective, we propose RestoreMore, which preserves the
pretrained restoration model as a frozen capability anchor and introduces
lightweight residual experts for newly arriving degradations.
A null expert retains the original restoration pathway when no additional
correction is required.
RestoreMore further employs capability-oriented bi-level routing at selected
feature stages. The first level identifies the Top-$M$ relevant restoration
capabilities, while the second selects and combines a sparse set of Top-$K$
complementary experts.
This stage-wise design enables each newly arriving task to selectively reuse
restoration knowledge accumulated from previous tasks without modifying the
pretrained backbone or historical experts.

The contributions of this work are summarized as follows:

\begin{itemize}
    \item We propose RestoreMore, a framework for continual restoration capability expansion that enables an already trained image restoration model to progressively acquire capabilities for previously unsupported degradations.

    \item We propose a capability-preserving expansion framework with
capability-oriented bi-level routing. The pretrained model is retained as a
frozen restoration anchor, while lightweight residual experts are incrementally
introduced for new degradations. Stage-wise capability retrieval and sparse
expert composition enable newly arriving tasks to selectively reuse
complementary restoration knowledge accumulated from previous tasks.

    \item Extensive experiments demonstrate that RestoreMore consistently acquires new restoration abilities while preserving and improving previously learned capabilities.
\end{itemize}

\section{Related Work}
\label{sec:related_work}

\subsection{Image Restoration}
\label{subsec:related_ir}

Image restoration aims to recover high-quality visual content from observations. 
Prior-driven restoration methods formulate image recovery using explicit
structural assumptions or degradation models~\cite{10558778,11342300}. 
Although these methods provide clear interpretations, their manually defined assumptions are often insufficient for complex and spatially varying degradations. 
The development of deep learning has shifted image restoration toward data-driven representation learning, allowing restoration models to learn degradation characteristics directly from large-scale datasets~\cite{guo2025mambairv2,FDTANetgao2025frequency}.

ALGNet~\cite{ALGgao2024learning} combines adaptive local and global feature extraction to recover both fine textures and large-scale structures, while XYScanNet~\cite{liu2024xyscannet} employs alternating spatial scans to improve long-range dependency modeling. 
EfDeRain+~\cite{efderainguo2025efficientderain+} treats deraining as a predictive filtering process rather than explicitly decomposing the rain layer. 
PGH$^2$Net~\cite{PGH2Netisu2025prior} incorporates bright- and dark-channel priors together with histogram equalization to guide hierarchical dehazing. 
Diffusion-based methods, including UPID-EDM~\cite{upid10.1145/3664647.3680560} and Diff-Unmix~\cite{diffunmix10656884}, further exploit generative priors and iterative denoising to improve perceptual fidelity. 
FSNet~\cite{FSNet} reorganize spatial and frequency information to improve generalization across restoration settings. 
ACL~\cite{aclgu2025acl} introduces efficient long-range modeling through linear attention, while StarIR~\cite{starir11429607} and MHNet~\cite{gao2025mixed} exploit hierarchical or mixed-frequency representations to capture restoration cues shared by different degradations. 
BaryIR~\cite{baryir11417902} aligns multiple degraded distributions through a barycentric representation, thereby improving degradation invariance and cross-task generalization. 
AdaIR~\cite{cui2025adair} disentangles degradation and content information in spatial and frequency domains, whereas Perceive-IR~\cite{Perceive-IR10990319} estimates both degradation categories and fine-grained severity to support more adaptive restoration. 
AllRestore~\cite{Allrestor11367271} combines visual and textual information into a composite scene embedding for describing degradation conditions within a unified model.
Large pretrained models and generative priors have further broadened the design space of all-in-one restoration. 
VLU-Net~\cite{VLUNetZeng_2025_CVPR} uses vision-language representations to extract degradation-aware cues and integrates them into an unfolding-based restoration process. 
AutoDIR~\cite{autodir10.1007/978-3-031-73661-2_19} combines vision-language guidance with latent diffusion priors, while Defusion~\cite{DefusionLuo_2025_CVPR} constructs degradation instructions and performs diffusion in the degradation space for generalized restoration. 

Despite these advances, existing restoration methods are still mainly developed under a fixed-capability assumption. 
When an unsupported degradation appears, the common solution is to train another model or retrain the unified model using both new and historical datasets.

\subsection{Continual Learning}
\label{subsec:related_cl}

Continual learning aims to enable a model to acquire knowledge from a sequence of tasks while mitigating catastrophic forgetting. Existing methods are commonly based on regularization, replay, optimization constraints, or architectural expansion. Regularization-based approaches restrict changes to parameters or predictions that are important for previously learned tasks~\cite{li2017learning}, whereas replay-based methods revisit stored or generated historical samples~\cite{rebuffi2017icarl,lopez2017gem}. 
The development of pretrained models has further motivated parameter-efficient continual learning. MoE-Adapters~\cite{yu2024moeadapters} dynamically introduces and selects adapter experts for continual vision-language learning. SEMA~\cite{wang2025sema} evaluates whether existing adapters can represent an incoming data distribution and expands the module set only when necessary. CaRE~\cite{lou2026care} further adopts a two-stage routing strategy, in which task-relevant routers are first selected and then used to retrieve and aggregate complementary experts at individual network layers. These methods demonstrate that modular adaptation and sparse knowledge retrieval can improve the stability--plasticity balance in continual learning with pretrained models.
Continual learning has also been extended to image restoration. KR~\cite{cheng2024continual} combines knowledge replay with feature-level distillation for sequential adverse-weather removal. ILAWR~\cite{lu2025continuous} employs degradation-aware distillation to reduce interference between newly introduced and previously learned weather-removal tasks. MINI~\cite{gong2025minimalistic} separates task-specific restoration knowledge through meta-convolutional parameters, thereby reducing destructive updates across incremental tasks.

Nevertheless, existing continual restoration methods mainly focus on
mitigating forgetting within restricted restoration task families, without
explicitly preserving an arbitrary pretrained restoration pathway or enabling
stage-wise reuse of complementary knowledge across degradation types.
In contrast, RestoreMore freezes the pretrained restorer as a capability
anchor and progressively expands its capability set through lightweight
residual experts and capability-oriented bi-level routing.

\section{Method}
\label{sec:method}

\subsection{Problem Formulation}
\label{subsec:problem}

We study continual restoration capability expansion, whose goal is to enable an already trained image restoration model to progressively restore degradation types that were not included in its original training set.
Let $F_0(\cdot;\theta_0)$ denote a pretrained image restoration model with an initial capability set $\mathcal{S}_0=\{d_1,d_2,\ldots,d_B\}$,
where each $d_i$ represents a degradation type supported by $F_0$. 
The initial model may be a task-specific restorer with $B=1$ or an all-in-one restorer with $B>1$.

After the initial training, previously unsupported degradation types arrive sequentially.
At incremental stage $t$, a paired dataset $\mathcal{D}_t ={(x_{t,n},y_{t,n})}_{n=1}^{N_t}$ is provided for a new degradation $d_{B+t}$, where $x_{t,n}$ and $y_{t,n}$ denote the degraded image and its clean target, respectively.
The capability set is then expanded as 
$\mathcal{S}_t = \mathcal{S}_{t-1} \cup \{d_{B+t}\}$.
Before continual expansion, a one-time capability indexing procedure is performed for $\mathcal{S}_0$.
Afterwards, historical training images are neither stored nor revisited, and each incremental stage uses only the newly arriving dataset $\mathcal{D}_t$.
Details of the initial capability indexing are provided in Appendix~\ref{app:init}.
The expanded model should restore all degradation types observed so far without requiring a degradation label
$\hat{y}=F_t(x)$, $d(x)\in\mathcal{S}_t$.
The continual expansion process considers two primary objectives:
\begin{equation}
\begin{aligned}
    \text{Acquisition:}\quad
    &Q_{t,B+t}\uparrow,\\
    \text{Retention:}\quad
    &Q_{t,i}\geq Q_{t-1,i}-\epsilon,
    \qquad d_i\in\mathcal{S}_{t-1},
\end{aligned}
\label{eq:continual_objective}
\end{equation}
where $Q_{t,i}$ denotes the restoration quality of $F_t$ on degradation $d_i$.
Unlike conventional all-in-one restoration, which jointly optimizes a model over a fixed collection of degradations, our formulation progressively expands the capability boundary of an existing restorer.
The pretrained backbone and previously learned expansion components remain frozen, while lightweight components are introduced only for newly arriving degradations.

\begin{figure}
    \centering
    \includegraphics[width=\linewidth]{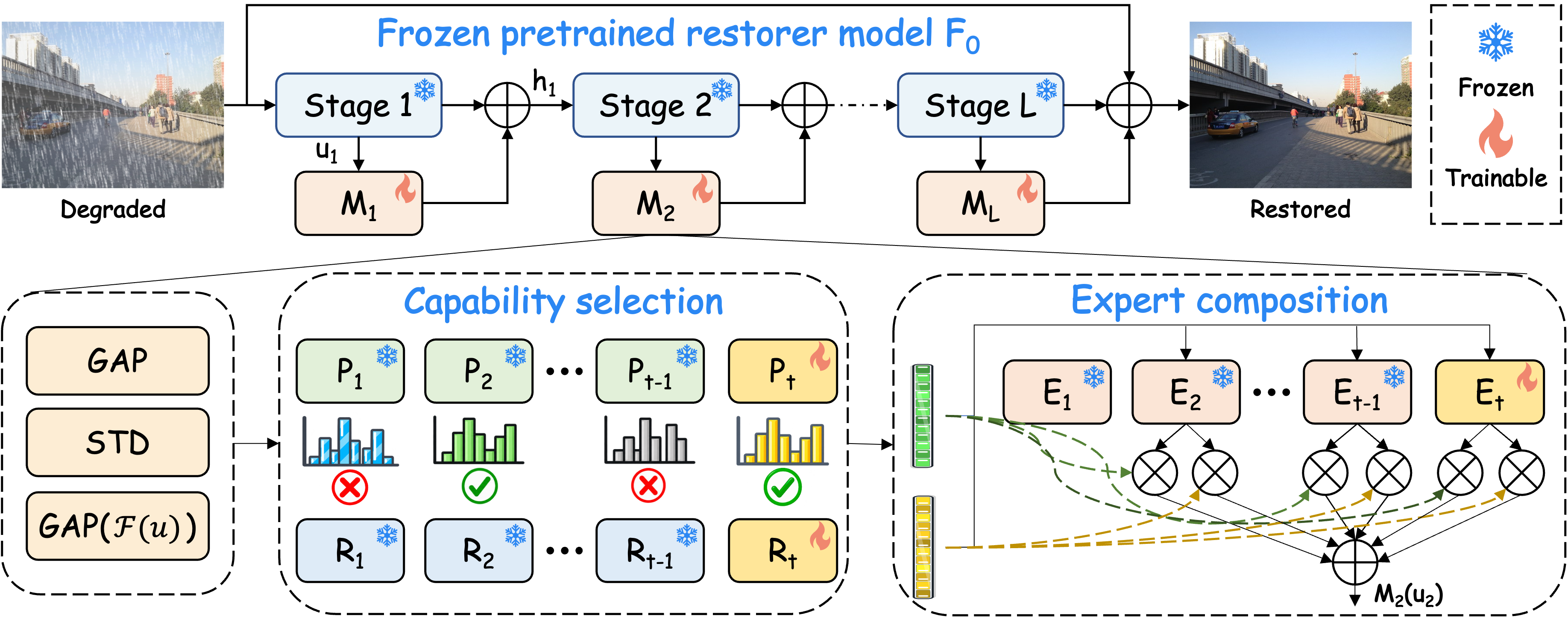}
    \caption{
    Overview of the proposed RestoreMore framework.
    A pretrained restoration model is preserved as a frozen capability anchor, and lightweight residual expansion modules are inserted at selected feature stages.
    Inside each expansion module, a degradation descriptor is first constructed from spatial and frequency statistics.
    Top-$M$ capability selection is then performed over accumulated capability prototypes and routers.
    The selected routers further retrieve complementary experts from the expert-key bank through query--key matching.
    Frozen historical components and trainable current-task components are indicated by snowflake and flame icons, respectively.
    }
    \label{fig:restoremore_framework}
\end{figure}

\subsection{Overall Pipeline}
\label{subsec:pipeline}
The overall pipeline of RestoreMore is illustrated in Fig.~\ref{fig:restoremore_framework}.
The pretrained restorer remains frozen and provides the original restoration pathway, while stage-wise residual expansion modules are introduced to acquire new degradation-specific capabilities.
Suppose that the pretrained restorer contains $L$ major feature-processing stages.
The original feature at stage $l$ is computed as
\begin{equation}
    u_l
    =
    \mathcal{B}_l
    \left(
        h_{l-1};\theta_{0,l}
    \right),
    \qquad
    l=1,\ldots,L,
    \label{eq:backbone_stage}
\end{equation}
where $\mathcal{B}_l$ denotes the original restoration transformation.
Its parameters remain frozen throughout continual expansion $ \theta_0^{(t)}=\theta_0,
 \forall t$. At selected stages, RestoreMore introduces a residual capability expansion module $\mathcal{M}_l^{(t)}$:
\begin{equation}
    h_l
    =
    u_l
    +
    \mathcal{M}_l^{(t)}(u_l).
    \label{eq:expanded_stage}
\end{equation}
Thus, the pretrained feature $u_l$ always remains in the forward pathway, while the expansion branch provides only the additional transformation required by incrementally acquired capabilities.
When a new degradation $d_{B+t}$ arrives, a capability prototype and router together with a degradation expert and its routing key are introduced at each selected feature stage.
All historical prototypes, routers, experts, and routing keys remain frozen.
Only the newly introduced components are optimized using $\mathcal{D}_t$.
Once training is completed, they are frozen and incorporated into the capability bank for subsequent tasks.
For notational simplicity, the aggregate effect of all stage-wise expansion modules can be abstractly expressed as:
\begin{equation}
    F_t(x)
    =
    F_0(x)
    +
    \Delta F_t
    \left(
        x;
        \mathcal{P}^{(t)},
        \mathcal{R}^{(t)},
        \mathcal{E}^{(t)}
    \right),
    \label{eq:expanded_model}
\end{equation}
where $\mathcal{P}^{(t)}$, $\mathcal{R}^{(t)}$, and $\mathcal{E}^{(t)}$ denote the accumulated capability prototypes, capability routers, and expert bank, respectively.
The complete incremental optimization procedure is given in Appendix~\ref{app:optimization}.

\subsection{Capability-Oriented Bi-Level Routing}
\label{subsec:routing}

The core of RestoreMore is a capability-oriented bi-level routing mechanism that selectively retrieves and composes restoration knowledge accumulated during continual learning.
At each feature stage, the first routing level identifies restoration capabilities relevant to the input, while the second selects complementary degradation experts associated with the retrieved capabilities.
At stage $l$, RestoreMore maintains ${(p_{i,l},R_{i,l})}_{d_i\in\mathcal{S}_t}$, $p_{i,l}$ and $R_{i,l}$ denote the prototype and router of capability $d_i$, respectively.
The incremental expert bank is:
\begin{equation}
    \mathcal{E}_l^{(t)}
    =
    \left\{
        (E_{\varnothing,l},k_{\varnothing,l}),
        (E_{1,l},k_{1,l}),
        \ldots,
        (E_{t,l},k_{t,l})
    \right\},
    \label{eq:expert_bank}
\end{equation}
where $k_{j,l}\in\mathbb{R}^{d_r}$ is the routing key associated with expert $E_{j,l}$.
The null expert satisfies $E_{\varnothing,l}(u)=0$, allowing the expansion branch to apply no additional correction when the pretrained pathway is sufficient.

\paragraph{Capability selection.}
Unlike recognition, image restoration does not naturally provide a discrete semantic representation for identifying the underlying restoration requirement.
We therefore construct a degradation-sensitive descriptor using spatial statistics and frequency characteristics:
\begin{equation}
    q_l
    =
    \psi_l
    \left(
        \left[
            \operatorname{GAP}(u_l),
            \operatorname{STD}(u_l),
            \operatorname{GAP}
            \left(
                |\mathcal{F}(u_l)|
            \right)
        \right]
    \right),
    \label{eq:degradation_descriptor}
\end{equation}
where $\operatorname{GAP}$ and $\operatorname{STD}$ denote channel-wise spatial average and standard deviation, $\mathcal{F}$ denotes the two-dimensional Fourier transform, and $\psi_l$ is a lightweight projection network.
To keep the descriptor space consistent across incremental stages, $\psi_l$ is calibrated during capability indexing and subsequently frozen; details are provided in Appendix~\ref{app:init}.
For each learned degradation $d_i$, the capability affinity $ a_{i,l} = \frac{q_l^{\top}p_{i,l}}{ \|q_l\|_2\|p_{i,l}\|_2 }$. 
The normalized capability probability is:
\begin{equation}
    \pi_{i,l}
    =
    \frac{
        \exp(a_{i,l}/\tau_c)
    }{
        \sum_{d_n\in\mathcal{S}_t}
        \exp(a_{n,l}/\tau_c)
    },
    \label{eq:capability_probability}
\end{equation}
where $\tau_c$ is a temperature parameter.
The first routing level selects the $M$ most relevant restoration capabilities
    $\mathcal{I}_l = \operatorname{TopM}( \{\pi_{i,l}\}_{d_i\in\mathcal{S}_t})$.
The selected probabilities are renormalized as:
\begin{equation}
    \widetilde{\pi}_{i,l}
    =
    \frac{
        \pi_{i,l}
    }{
        \sum_{n\in\mathcal{I}_l}\pi_{n,l}
    }.
\end{equation}

Selecting multiple capabilities enables the model to exploit complementary restoration knowledge.
For example, rain and snow removal may share high-frequency suppression patterns, while denoising and deblurring can provide complementary texture and structural cues.
Stage-wise routing further allows different feature depths to retrieve different forms of historical knowledge.

\paragraph{Expert composition.}
For each selected capability, the corresponding router maps $q_l$ to a fixed-dimensional routing query $ r_{i,l} = R_{i,l}(q_l) \in \mathbb{R}^{d_r}$. The compatibility between capability router $i$ and expert $j$ is computed through query--key matching:
\begin{equation}
    z_{ij,l}
    =
    \frac{
        r_{i,l}^{\top}k_{j,l}
    }{
        \|r_{i,l}\|_2
        \|k_{j,l}\|_2
    },
    \label{eq:expert_affinity}
\end{equation}
\begin{equation}
    g_{ij,l}
    =
    \frac{
        \exp(z_{ij,l}/\tau_e)
    }{
        \sum_{n\in\{\varnothing,1,\ldots,t\}}
        \exp(z_{in,l}/\tau_e)
    }.
    \label{eq:expert_probability}
\end{equation}

Because $r_{i,l}$ has a fixed dimensionality, adding a new expert requires only appending an expert--key pair $(E_{t,l},k_{t,l})$ and does not modify the structure or parameters of historical routers.
A detailed explanation of this property is provided in Appendix~\ref{app:querykey}.
Only the Top-$K$ experts are activated for each selected capability:
\begin{equation}
    \mathcal{J}_{i,l}
    =
    \operatorname{TopK}
    \left(
        \{g_{ij,l}\}_{j\in\{\varnothing,1,\ldots,t\}}
    \right).
    \label{eq:topk_experts}
\end{equation}
Their outputs are combined as:
\begin{equation}
    v_{i,l}
    =
    \sum_{j\in\mathcal{J}_{i,l}}
    \widetilde{g}_{ij,l}
    E_{j,l}(u_l),
    \label{eq:expert_composition}
\end{equation}
where $\widetilde{g}_{ij,l}$ denotes the expert weights renormalized over $\mathcal{J}_{i,l}$.
The final expansion residual is $\mathcal{M}_l^{(t)}(u_l)
    =
    \sum_{i\in\mathcal{I}_l}
    \widetilde{\pi}_{i,l}
    v_{i,l}$. Therefore, only a sparse subset of accumulated experts participates in each forward pass.
The expert architecture, cold-start strategy, optimization objective, and backbone-specific implementation are deferred to Appendix~\ref{app:optimization}--\ref{app:instantiation}, while the computational complexity is discussed in Appendix~\ref{app:complexity}.

\section{Experiments}
\label{sec:exp}
In this section, we first present the qualitative and quantitative comparison results, followed by the ablation studies. The experimental setup and more experiments are  provided in Appendix~\ref{sec:expsetup} and~\ref{sec:expset}.

\begin{table*}
\centering
\caption{Quantitative comparison of different capability-expansion strategies on CNN-, Transformer-, and Mamba-based image restoration models. 
$^{O}$, $^{A}$, and $^{R}$ denote the original pretrained model, joint retraining with mixed old and new datasets, and RestoreMore trained only with newly arriving task data, respectively.}
\label{tb:allrshbn}
    \resizebox{\linewidth}{!}{
\begin{tabular}{c|cccccccccc|cc}
    \hline
     \multirow{2}{*}{Methods} & \multicolumn{2}{c}{Deraining} & \multicolumn{2}{c}{Desnowing} & \multicolumn{2}{c}{Dehazing}& \multicolumn{2}{c}{Deblurring}& \multicolumn{2}{c|}{Denoising}  & \multicolumn{2}{c}{Average} 
    \\
    &PSNR $\uparrow$ &  SSIM $\uparrow$  &PSNR $\uparrow$ &SSIM $\uparrow$ & PSNR $\uparrow$&SSIM $\uparrow$ &PSNR $\uparrow$ & SSIM $\uparrow$&PSNR $\uparrow$ & SSIM $\uparrow$&PSNR $\uparrow$ & SSIM $\uparrow$
    \\
    \hline\hline
     ALGNet$^{O}$~\cite{ALGgao2024learning} & - & - &- &- & - &-  &33.49 &0.964 &- &- &- &-
     \\
     ALGNet$^{A}$~\cite{ALGgao2024learning}& 35.89 & 0.969 & 30.89 & 0.924 & 32.76 & 0.968 & 29.41 & 0.882 & 30.97 & 0.881 & 31.98 & 0.925
     \\
    \rowcolor{gray!20} \textbf{ALGNet$^{R}$}~\cite{ALGgao2024learning}& 36.37 & 0.973 & 30.97 & 0.927 & 34.47 & 0.974 & 33.18 & 0.952 & 31.31 & 0.885 &33.26  &0.942 
     \\
     \hline
     Perceive-IR$^{O}$~\cite{Perceive-IR10990319}& 38.29 & 0.980 & - & - & 30.87 & 0.975 & 29.46 & 0.886 & 31.53 & 0.890 & - & -
     \\
      Perceive-IR$^{A}$~\cite{Perceive-IR10990319}& 38.92 & 0.981 & 32.99 & 0.948 & 34.52 & 0.980 & 29.79 & 0.889 & 31.44 & 0.885 & 33.53 & 0.937
      \\
      \rowcolor{gray!20} \textbf{Perceive-IR$^{R}$}~\cite{Perceive-IR10990319}& 39.25 & 0.982 & 33.15 & 0.949 & 34.77 & 0.981 & 31.73 & 0.927 & 31.62 & 0.891 & 34.10 & 0.946
        \\
      \hline
 StarIR$^{O}$~\cite{starir11429607} &38.50 &0.984 & -& - & 30.89 &0.979 & - &-& 31.51 & 0.893 & -& -
      \\
        StarIR$^{A}$~\cite{starir11429607} &38.71 &0.981 & 33.15& 0.950 & 34.55&0.978 & 29.82 &0.887& 31.45 & 0.885 & 33.54 & 0.936
        \\ 
        \rowcolor{gray!20} \textbf{StarIR$^{R}$}~\cite{starir11429607} &39.42 &0.983 & 33.26& 0.952 & 34.80 &0.982 & 31.63 &0.928& 31.58 & 0.889 &34.14  &0.947 
    \\
    \hline
\end{tabular}}
\end{table*}

\begin{figure*} 
    \centerline{\includegraphics[width=1\linewidth]{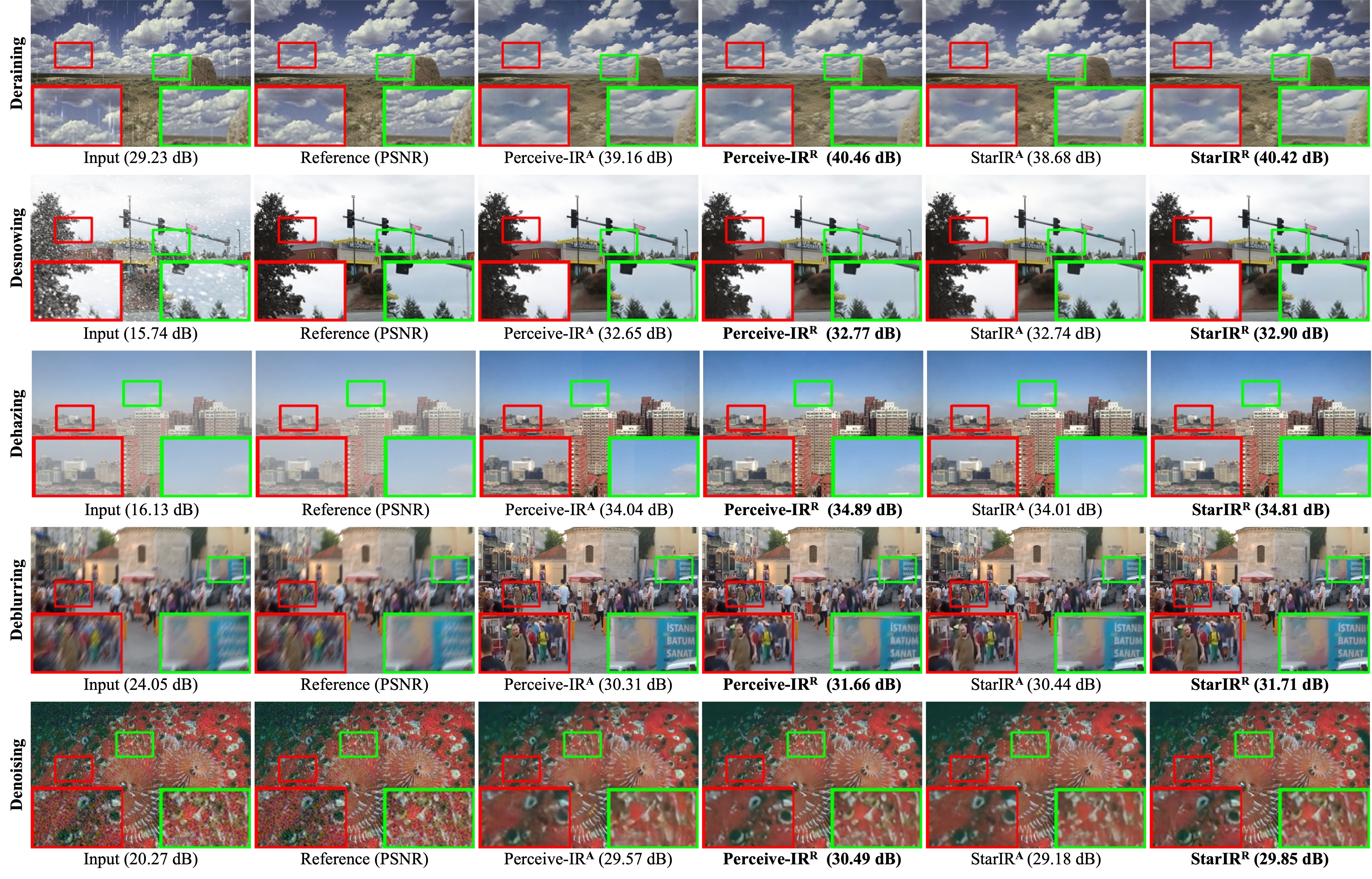}}
	\caption{Qualitative comparison across five restoration tasks.}
 \label{fig:qualitative}
\end{figure*}

\subsection{Results}

Table~\ref{tb:allrshbn} evaluates RestoreMore on representative CNN- (StarIR~\cite{starir11429607}), Transformer- (Perceive-IR~\cite{Perceive-IR10990319}), and Mamba-based (ALGNet~\cite{ALGgao2024learning}) restoration models by comparing three settings: the original pretrained model ($^O$), joint retraining with mixed old and new datasets ($^A$), and continual capability expansion with RestoreMore ($^R$). 
Despite using only the datasets of newly arriving tasks rather than revisiting the original training data, RestoreMore consistently outperforms joint retraining across all three architectures. 
Compared with the all-in-one retraining baseline, RestoreMore improves the average PSNR/SSIM from 31.98/0.925 to 33.26/0.942 for ALGNet, from 33.53/0.937 to 34.10/0.946 for Perceive-IR, and from 33.54/0.936 to 34.14/0.947 for StarIR. 
These consistent gains indicate that the proposed capability-expansion mechanism is not tied to a particular restoration architecture.

The results also show that RestoreMore better preserves the capabilities of the original models. 
This is particularly evident for ALGNet, whose original deblurring performance is 33.49 dB, but decreases substantially to 29.41 dB after mixed-data retraining. 
RestoreMore retains 33.18 dB while simultaneously acquiring four additional restoration capabilities, reducing the degradation on the original task from 4.08 dB to only 0.31 dB. 
For the stronger multi-task pretrained models, RestoreMore not only maintains historical performance but can further improve it. 
For example, Perceive-IR$^{R}$ improves the original results by 0.96 dB on deraining, 3.90 dB on dehazing, and 2.27 dB on deblurring. 
These results show that preserving the original restoration backbone while
selectively reusing accumulated experts can effectively retain historical
capabilities during continual expansion. Interestingly, several historical
tasks even outperform their original models, suggesting that the expanded
expert bank can provide complementary restoration knowledge beyond the
initial capability set.

Meanwhile, RestoreMore remains effective on the newly introduced tasks. 
For ALGNet, which initially supports only deblurring, RestoreMore achieves higher performance than joint retraining on all four newly introduced tasks, with particularly clear gains of 1.71 dB on dehazing. 
Similar improvements are observed for Perceive-IR and StarIR, including gains of 1.94 dB and 1.81 dB on deblurring over their corresponding joint-training baselines.

Fig.~\ref{fig:qualitative} provides qualitative comparisons across the five restoration tasks.
Consistent with the quantitative results in Table~\ref{tb:allrshbn}, RestoreMore produces visually cleaner results while retaining more faithful structures and local details.
For deraining and desnowing, residual rain streaks and snow-like artifacts are more effectively suppressed without noticeably degrading the background content.
For dehazing, RestoreMore recovers clearer scene visibility and more natural contrast, particularly around distant buildings and sky regions.
In the deblurring examples, object boundaries, fine structures, and text regions are reconstructed more sharply, while competing results tend to retain blur or lose local details.
For denoising, RestoreMore suppresses noise while better preserving fine textures instead of introducing excessive smoothing.

\begin{table*}
\centering
\caption{Zero-shot generalization results on real-world degradation datasets.}
\label{tb:allrealw}
\resizebox{\linewidth}{!}{
\begin{tabular}{c|cccccccc}
    \hline
    \multirow{2}{*}{Methods} & \multicolumn{3}{c}{RealRain-1k-L} & \multicolumn{3}{c}{RTTS} & \multicolumn{2}{c}{SIDD}
    \\
    &PSNR $\uparrow$ &  SSIM $\uparrow$ &LPIPS $\downarrow$ &FADE $\downarrow$ &BRISQUE $\downarrow$ &NIMA $\uparrow$&PSNR $\uparrow$ & SSIM $\uparrow$
    \\
    \hline\hline
   
  ECFNet~\cite{gao2026emphasizing} & 23.69 & 0.756 & 0.401 & 1.394 & 29.105 &4.372 & 24.33 & 0.471 
    \\
   
        Defusion~\cite{DefusionLuo_2025_CVPR}& \underline{27.37} & \textbf{0.903} & \underline{0.371} & 1.277 & \underline{23.520} & \underline{4.615} & 24.52 & 0.495
        \\
         Perceive-IR~\cite{Perceive-IR10990319}& 27.31 & \underline{0.901}  & 0.372 & \underline{1.264} & 23.694 &4.613 & \underline{24.53} & \underline{0.497}
        \\
      \rowcolor{gray!20} \textbf{Perceive-IR$^R$}~\cite{Perceive-IR10990319} & \textbf{27.55} & \textbf{0.903}  & \textbf{0.369} & \textbf{1.240} & \textbf{23.494} &\textbf{4.616} & \textbf{24.61} & \textbf{0.499}
    \\
    \hline
\end{tabular}}
\end{table*}

\subsection{Generalization}

\subsubsection{Zero-shot Generalization to Real-World Scenes}

As shown in Table~\ref{tb:allrealw}, applying RestoreMore to Perceive-IR consistently improves its real-world generalization across deraining, dehazing, and denoising. 
On RealRain-1k-L~\cite{realrain1li2022toward}, Perceive-IR$^R$ achieves the best PSNR and LPIPS of 27.55~dB and 0.369, respectively, while obtaining the best SSIM of 0.903 jointly with Defusion. 
Compared with the original Perceive-IR, RestoreMore improves PSNR by 0.24~dB and reduces LPIPS from 0.372 to 0.369.
Similar improvements are observed on real-world hazy and noisy images.  
These consistent improvements indicate that continual capability expansion does not compromise the generalization ability of the pretrained model. Instead, the newly acquired restoration knowledge can improve robustness to real-world degradation distributions.

\begin{table}
\centering
\caption{Zero-shot generalization results on an unseen degradation type.}
\label{tb:allundt}
\begin{tabular}{c|ccccc}
    \hline
    \multirow{2}{*}{Methods} & \multicolumn{3}{c}{UIEB} & \multicolumn{2}{c}{C60}
    \\
   & PSNR $\uparrow$ &  SSIM $\uparrow$ &LPIPS $\downarrow$ &UCIQE$\uparrow$ &UIQM$\uparrow$
    \\
    \hline\hline
ECFNet~\cite{gao2026emphasizing} & 20.49 & 0.856 & 0.223 & 0.539 & 2.517 
    \\
ACL~\cite{aclgu2025acl} & 20.94 & 0.867 & 0.201 & 0.552 & 2.485 
     \\

      Perceive-IR~\cite{Perceive-IR10990319}& \underline{21.77} &\underline{0.891} & \underline{0.169} & \underline{0.557} & \underline{2.555 }
        \\
      \rowcolor{gray!20}\textbf{Perceive-IR$^{R}$}~\cite{Perceive-IR10990319} & \textbf{21.92} & \textbf{0.894} & \textbf{0.165} & \textbf{0.558} & \textbf{2.559}
    \\
    \hline
\end{tabular}
\end{table}

\subsubsection{Generalization to an Unseen Degradation Type}
We further evaluate whether the expanded model can generalize beyond the degradation types observed during training. 
As reported in Table~\ref{tb:allundt}, Perceive-IR$^R$ achieves the best results across all metrics. 
Although the improvements are moderate, their consistency across both full-reference and no-reference metrics suggests that continual capability expansion preserves, and can slightly enhance, the pretrained model's ability to generalize to degradation types never encountered during training.

\begin{table}
\centering
\caption{Zero-shot generalization results on unseen noise levels.}
\label{tb:allunds}
\begin{tabular}{c|cccc}
    \hline
    \multirow{2}{*}{Methods} & \multicolumn{2}{c}{CBSD68} & \multicolumn{2}{c}{Urban100}
    \\
   & 60 & 100 & 60 & 100
    \\
    \hline\hline
     VLU-Net~\cite{VLUNetZeng_2025_CVPR}&  27.11 & 20.59 & 27.64 & 21.53
      \\
      Perceive-IR~\cite{Perceive-IR10990319}& \underline{27.13} & 20.65 & 27.65 & \underline{21.55}
        \\
        Defusion~\cite{DefusionLuo_2025_CVPR}& 27.11 & \underline{20.72} & \textbf{27.67} & 21.49
        \\
      \rowcolor{gray!20}\textbf{Perceive-IR$^R$}~\cite{Perceive-IR10990319} & \textbf{27.27} & \textbf{20.74} & \underline{27.66} & \textbf{21.58}
    \\
    \hline
\end{tabular}
\end{table}

\subsubsection{Generalization to Unseen Degradation Severity}
We additionally examine generalization to degradation severities outside the training distribution. 
The denoising task is trained with noise levels $\sigma\in\{15,25,50\}$, while evaluation is conducted at the unseen and substantially stronger noise levels of $\sigma=60$ and $100$.
As shown in Table~\ref{tb:allunds}, Perceive-IR$^R$ achieves the best performance in three of the four settings and remains within 0.01~dB of the best result on Urban100 at $\sigma=60$.  
The gains remain consistent as the noise level moves beyond the training range, indicating that the additional capabilities introduced by RestoreMore do not lead to over-specialization to the observed degradation severities.

\subsection{Ablation Studies}
\subsubsection{Effect of Each Component}

\begin{table}
    \centering
    \caption{Ablation study on individual components of RestoreMore.}
    \label{tab:abl1}
    \begin{tabular}{cccccc}
    \hline
        Ssettings& Frozen & Capability routing& Expert routing &  PSNR &$\triangle$ PSNR 
         \\
         \hline
         \hline
         (a)  & & & & 31.98 & -
         \\
         (b)  &\ding{52}&\ding{52} &  & 32.56 & + 0.58 dB
         \\
          (c)  &\ding{52} & &\ding{52}  & 32.69 & + 0.71 dB
         \\
        (d)  & &\ding{52} &\ding{52}  & 32.25 & + 0.27 dB
         \\
          (e)  &\ding{52} &\ding{52} &\ding{52}  & 33.26 & + 1.28 dB
        \\
         \hline
    \end{tabular}
\end{table}

We investigate the contribution of the main components of RestoreMore using ALGNet as the base restoration model. 
As shown in Table~\ref{tab:abl1}, setting (a) retrains ALGNet on the mixed datasets of all degradation types and serves as the baseline, achieving an average PSNR of 31.98~dB. 
In contrast, settings (b)--(e) use only the dataset of the newly arriving degradation during continual expansion. 
Despite the absence of historical raw data, all variants outperform the jointly retrained baseline, demonstrating the effectiveness of expanding restoration capabilities from newly available task data.

Freezing the pretrained restoration model plays an important role in preserving previously learned mappings. 
When the frozen capability anchor is combined with capability routing or expert routing, settings (b) and (c) improve the baseline by 0.58~dB and 0.71~dB, respectively. 
Expert routing provides a slightly larger gain, suggesting that selectively reusing complementary historical experts is particularly beneficial for acquiring new restoration capabilities. 
By comparison, jointly applying capability and expert routing without freezing the original model in setting (d) yields only a 0.27~dB improvement. 
This result indicates that routing alone cannot fully prevent interference with previously learned restoration knowledge when the backbone remains trainable.
The complete RestoreMore in setting (e) combines the frozen capability anchor with both levels of routing and achieves the best average PSNR of 33.26~dB, outperforming the mixed-data retraining baseline by 1.28~dB. 
It also improves upon the strongest partial variant (c) by 0.57~dB. 
These results show that the three components are complementary.

\begin{table}
    \centering
    \caption{Effects of the order in which new restoration tasks are learned.}
    \label{tab:ablseq}
    \begin{tabular}{c|cccc}
    \hline
        Method& R $\rightarrow$ S $\rightarrow$ H $\rightarrow$ N &  N $\rightarrow$ R $\rightarrow$ S $\rightarrow$ H &   H $\rightarrow$ N  $\rightarrow$ R $\rightarrow$ S &  S $\rightarrow$  H   $\rightarrow$ R $\rightarrow$ N 
         \\
         \hline
         \hline
         PSNR & 33.26 & 33.19 & 33.24 & 33.20
         \\
         \hline
    \end{tabular}
\end{table}

\subsubsection{Effect of Task Order}
We further evaluate the sensitivity of RestoreMore to the order in which new restoration tasks are introduced. 
As shown in Table~\ref{tab:ablseq}, the average PSNR remains highly stable across four different task sequences. 
This small variation indicates that RestoreMore is not strongly dependent on a specific task order. 
The frozen capability anchor limits interference with previously learned restoration mappings, while capability-oriented routing and expert reuse allow newly arriving tasks to selectively exploit relevant historical knowledge regardless of when they are introduced.

\section{Conclusion}
\label{sec:conclusion}

In this work, we  propose RestoreMore, which preserves the pretrained restorer as a frozen capability anchor and incrementally introduces lightweight residual expansion modules for newly arriving degradations. 
A capability-oriented bi-level routing mechanism was developed to first identify relevant restoration capabilities and then sparsely select complementary degradation experts at multiple feature stages. 
This design enables effective reuse of accumulated restoration knowledge for
new tasks while progressively enriching the expert bank for subsequent
capability expansion.
Extensive experiments across diverse restoration benchmarks and model architectures demonstrate that RestoreMore consistently acquires new restoration abilities while preserving, and in many cases improving, existing ones.

\section*{AI Use Statement}

In this work, we used generative AI tools to assist with manuscript editing and
scientific figure preparation. GPT-based tools were used to improve the
language, grammar, and readability of the manuscript without changing the
underlying scientific content. Generative AI was additionally used to assist in
the preparation of Fig.~5, using only the experimental values obtained by the
authors as input. The underlying data and numerical results in the figure were
not generated or modified by AI.

All AI-assisted outputs were manually reviewed and verified by the authors.
The authors take full responsibility for the final manuscript, including all
text, figures, experimental results, scientific claims, and conclusions.

\bibliography{iclr2027_conference}
\bibliographystyle{iclr2027_conference}

\appendix
\section{Appendix}
\label{sec:appendix_method}

\subsection{Initialization of the Capability Bank}
\label{app:init}

RestoreMore starts from a pretrained model $F_0$ with initial capability set $\mathcal{S}_0$.
Before continual expansion, we perform a one-time indexing procedure to construct the stage-wise capability representations required by the first routing level.
This procedure is performed only for the initial capability set and does not update the pretrained restoration backbone.

For an initial degradation $d_i\in\mathcal{S}_0$, its samples are forwarded through $F_0$ and the degradation descriptor $q_l$ is extracted at each selected feature stage.
The projection $\psi_l$ is calibrated during this stage and is subsequently frozen so that descriptors from later tasks remain in the same representation space.
The initial capability prototype is computed as:
\begin{equation}
    p_{i,l}
    =
    \operatorname{Norm}
    \left[
        \frac{1}{|\mathcal{D}_i|}
        \sum_{x\in\mathcal{D}_i}
        q_l(x)
    \right].
    \label{eq:initial_prototype}
\end{equation}
The capability router $R_{i,l}$ is initialized jointly during this indexing stage.
Since the corresponding restoration capability is already represented by $F_0$, the null expert provides the default residual option for these initial capabilities.
For a newly arriving degradation $d_{B+t}$, its prototype is initialized from the current training features and updated using exponential moving averaging:
\begin{equation}
    p_{B+t,l}
    \leftarrow
    \eta p_{B+t,l}
    +
    (1-\eta)
    \frac{1}{|\mathcal{B}_t|}
    \sum_{x\in\mathcal{B}_t}
    q_l(x),
    \label{eq:prototype_ema}
\end{equation}
where $\mathcal{B}_t\subset\mathcal{D}_t$ denotes the current mini-batch.
Once stage $t$ is completed, the prototype is frozen.
After the initial indexing stage, historical raw training images are discarded.
All subsequent incremental stages therefore use only the dataset of the newly arriving degradation.

\subsection{Why Query--Key Routing Supports Continual Expansion}
\label{app:querykey}

A standard MoE router usually predicts one gating logit for each available expert.
Its output dimension therefore grows with the expert bank $ R_{i,l}(q_l) \in \mathbb{R}^{|\mathcal{E}_l^{(t)}|}$.
Adding a new expert would consequently require expanding the output layer of historical routers, which conflicts with our objective of keeping previously learned components unchanged.
RestoreMore instead separates capability routing from expert indexing.
Each capability router produces a fixed-dimensional query $ r_{i,l}\in\mathbb{R}^{d_r}$, while every expert owns a routing key $  k_{j,l}\in\mathbb{R}^{d_r}$. Their compatibility is evaluated through Eq.~\eqref{eq:expert_affinity}.
Consequently, adding a new capability only appends $(E_{t,l},k_{t,l})$ to the expert bank, without changing any historical router.

This design separates knowledge accumulation from historical parameter preservation.
New restoration knowledge is accumulated through an expanding expert--key bank, whereas previously learned routers and experts remain structurally unchanged.
The same query--key space also allows newly introduced experts to become candidates for future restoration tasks.

\subsection{Optimization and Cold-Start Strategy}
\label{app:optimization}

At incremental stage $t$, the pretrained backbone and all historical capability components are frozen.
The trainable parameters are:
\begin{equation}
    \Theta_t
    =
    \left\{
        R_{B+t,l},
        E_{t,l},
        k_{t,l}
    \right\}_{l=1}^{L},
\end{equation}
while the prototype $p_{B+t,l}$ is updated using Eq.~\eqref{eq:prototype_ema}.

For a training pair $(x_t,y_t)\in\mathcal{D}_t$, the restoration objective is:
\begin{equation}
\begin{split}
    \mathcal{L}_{\mathrm{res}}
    =
    &
    \sqrt{
        \|F_t(x_t)-y_t\|_2^2+\epsilon^2
    }\\
    &+
    \lambda_f
    \left\|
        \mathcal{F}(F_t(x_t))
        -
        \mathcal{F}(y_t)
    \right\|_1.
\end{split}
\label{eq:restoration_loss}
\end{equation}

During training, the current capability is always included in the Top-$M$ capability set:
\begin{equation}
    \mathcal{I}_l^{\mathrm{train}}
    =
    \{B+t\}
    \cup
    \operatorname{Top}_{M-1}
    \left(
        \{\pi_{i,l}\}_{i\neq B+t}
    \right).
    \label{eq:forced_capability}
\end{equation}
Similarly, the newly introduced expert is forced into the Top-$K$ set of its own capability router:
\begin{equation}
    \mathcal{J}_{B+t,l}^{\mathrm{train}}
    =
    \{t\}
    \cup
    \operatorname{Top}_{K-1}
    \left(
        \{g_{B+t,j,l}\}_{j\neq t}
    \right).
    \label{eq:forced_expert}
\end{equation}

This cold-start strategy guarantees that the newly introduced router, expert, and routing key receive sufficient optimization before their routing scores become reliable.
Meanwhile, the remaining $M-1$ capabilities and $K-1$ experts are dynamically selected from previously learned components, allowing the new task to reuse historical restoration knowledge from the beginning of training.
No historical image replay or output-level knowledge distillation is used.
After optimization, the new prototype, router, expert, and routing key are frozen and added to the capability bank.

\subsection{Expert Architecture and Backbone Instantiation}
\label{app:instantiation}

Each degradation expert is implemented as a lightweight bottleneck spatial adapter:
\begin{equation}
    E_{j,l}(u)
    =
    \gamma_{j,l}
    W_{j,l}^{\mathrm{up}}
    \left[
        \operatorname{DWConv}_{3\times3}
        \left(
            \sigma
            \left(
                W_{j,l}^{\mathrm{down}}
                \operatorname{Norm}(u)
            \right)
        \right)
    \right],
    \label{eq:expert_architecture}
\end{equation}
where $W_{j,l}^{\mathrm{down}}$ reduces the channel dimension,
$\operatorname{DWConv}_{3\times3}$ captures local spatial degradation patterns,
and $W_{j,l}^{\mathrm{up}}$ restores the original feature dimension.
The residual scale $\gamma_{j,l}$ is initialized to zero.
RestoreMore is attached to the outputs of major feature-processing stages rather than every individual block.
The same residual expansion interface $  h_l =u_l +\mathcal{M}_l^{(t)}(u_l)$ is used for CNN-, Transformer-, and Mamba-based restorers. Therefore, RestoreMore does not modify the internal convolution, attention, or state-space operations of the original backbone.

\subsection{Sparse Expert Computation}
\label{app:complexity}

Assume that $t$ incremental experts have been accumulated at each selected stage and that one expert requires computation $C_E$.
A dense expert mixture evaluates all experts and incurs $\mathcal{O}(tC_E)$ expert computation.
RestoreMore selects $M$ capability routers, each activating at most $K$ experts.
The dominant expert computation is therefore bounded by $\mathcal{O}(MKC_E)$.
Because $M$ and $K$ are fixed, the number of evaluated spatial experts does not grow with the task sequence.

Capability-prototype matching and query--key scoring still increase with the size of the capability bank.
However, these operations involve only low-dimensional vector similarities and are substantially cheaper than evaluating spatial restoration experts.
Thus, RestoreMore maintains sparse expert computation while progressively enlarging its available restoration knowledge.

\subsection{Experimental Setup}
\label{sec:expsetup}

\subsubsection{Datasets}
\textbf{i) Image deraining.}
For image deraining, we use 13,712 paired clean--rain images collected from multiple datasets~\cite{Rain100,Test100,8099669,7780668} for training. 
Evaluation is conducted on four commonly used benchmarks, including Rain100H~\cite{Rain100}, Rain100L~\cite{Rain100}, Test100~\cite{Test100}, and Test1200~\cite{DIDMDN}.

\textbf{ii) Image desnowing.}
For image desnowing, we adopt Snow100K~\cite{desnownet}, SRRS~\cite{JSTASRchen2020jstasr}, and CSD~\cite{HDCW-Netchen2021all}. 
Following the training protocol of previous work~\cite{FSNet}, 2,500 paired images are randomly sampled for training and 2,000 images are used for evaluation.

\textbf{iii) Image dehazing.}
For image dehazing, we use the daytime synthetic subsets of RESIDE~\cite{RESIDEli2018benchmarking}, including the Indoor Training Set (ITS), Outdoor Training Set (OTS), and Synthetic Objective Testing Set (SOTS). 
The model is trained separately on ITS and OTS and evaluated on SOTS-Indoor and SOTS-Outdoor, respectively, each containing 500 paired images.

\textbf{iv) Image deblurring.}
For image deblurring, we use the GoPro dataset~\cite{Gopro}, which contains 2,103 training pairs and 1,111 testing pairs.

\textbf{v) Image denoising.}
For image denoising, we construct a composite training set containing 800 images from DIV2K~\cite{DIK}, 2,650 images from Flickr2K~\cite{lim2017enhanced}, 400 images from BSD500~\cite{BSD500}, and 4,744 images from WED~\cite{ma2016waterloo}. 
Additive white Gaussian noise is synthesized with the noise level $\sigma$ randomly selected from $\{15,25,50\}$. 
The trained model is evaluated on CBSD68~\cite{BSD68}, Urban100~\cite{urban100}, and Kodak24~\cite{kodak}.

For the joint-retraining baseline, the training sets of the above restoration tasks are combined into a mixed dataset. 
For the unified quantitative comparison, we report results on Rain100L~\cite{Rain100} for deraining, Snow100K~\cite{desnownet} for desnowing, SOTS-Outdoor~\cite{RESIDEli2018benchmarking} for dehazing, GoPro~\cite{Gopro} for deblurring, and CBSD68~\cite{BSD68} with $\sigma=25$ for denoising.

\subsubsection{Training Details}
All models are optimized using Adam~\cite{2014Adam} with $\beta_1=0.9$ and $\beta_2=0.999$. 
The initial learning rate is set to $2\times10^{-4}$ and gradually decayed to $1\times10^{-7}$ using cosine annealing~\cite{2016SGDR}. 
During training, $256\times256$ image patches are randomly cropped with a batch size of 32, and each model is optimized for $4\times10^{5}$ iterations. 
Random horizontal and vertical flipping is adopted for data augmentation. 
For fair comparison, all deep learning-based baselines are fine-tuned or retrained following the hyperparameter settings reported in their original papers.

\subsubsection{Initial and Incrementally Acquired Restoration Capabilities}
\label{app:capability_config}

We evaluate RestoreMore using pretrained restoration models built on different architectural families.
Their original restoration capabilities are directly inherited from the corresponding pretrained models, while RestoreMore is used only to learn degradation types that are not initially supported.

We use the abbreviations R, S, H, B, and N to denote deraining, desnowing, dehazing, deblurring, and denoising, respectively.
The capability configurations used in our experiments are summarized in Table~\ref{tab:capability_configuration}. ALGNet initially provides only image deblurring capability.
We therefore use it to examine the more challenging setting in which a task-specific pretrained model is progressively expanded toward multi-degradation restoration.
Starting from B, RestoreMore sequentially learns four additional capabilities, R, S, H, and N, without revisiting the original deblurring training data.
Perceive-IR already supports R, H, B, and N in its pretrained form.
RestoreMore is applied only to the previously unsupported desnowing task, allowing us to evaluate whether an existing all-in-one model can be extended beyond its original task boundary.
StarIR initially supports R, H, and N.
RestoreMore further introduces S and B, providing another all-in-one-to-more setting with multiple incremental expansion stages.
These configurations together evaluate both task-specific-to-more and all-in-one-to-more capability expansion within the same framework.

For experiments involving multiple newly introduced tasks, the exact incremental orders are specified together with the corresponding experimental results.
In particular, ALGNet uses B as its initial capability, while R$\rightarrow$S$\rightarrow$H$\rightarrow$N is used as the default incremental sequence unless otherwise stated.

\begin{table}[t]
\centering
\caption{Initial and newly acquired restoration capabilities of the pretrained models used in our experiments. R, S, H, B, and N denote deraining, desnowing, dehazing, deblurring, and denoising, respectively.}
\label{tab:capability_configuration}
\resizebox{\linewidth}{!}{
\begin{tabular}{c|c|c|c}
\hline
Pretrained model
& Architecture
& Initial capabilities
& Newly acquired capabilities \\
\hline\hline
ALGNet~\cite{ALGgao2024learning}
& Mamba
& B
& R, S, H, N
\\
Perceive-IR~\cite{Perceive-IR10990319}
& Transformer
& R, H, B, N
& S
\\
StarIR~\cite{starir11429607}
& CNN
& R, H, N
& S, B
\\
\hline
\end{tabular}}
\end{table}

\subsubsection{Default Hyperparameters}
\label{app:hyperparameters}

Unless otherwise specified, capability expansion modules are inserted at the outputs of the major feature stages of each pretrained restorer.
We use $ M=2$, $ K=3$ for capability and expert selection, respectively.
When fewer than $K$ experts are available during early incremental stages, all available experts are considered. The routing temperatures are set to $\tau_c=0.1$, $\tau_e=1.0$, and the frequency-domain reconstruction weight is set to $\lambda_f=0.1$.
The same RestoreMore configuration is used across CNN-, Transformer-, and Mamba-based restorers unless otherwise specified.

\subsubsection{Evaluation Metrics} We evaluate performance using both reference-based and no-reference metrics. The reference-based metrics include Peak Signal-to-Noise Ratio (PSNR), Structural Similarity Index (SSIM), and Learned Perceptual Image Patch Similarity (LPIPS)~\cite{54zhang2018unreasonable}. The no-reference metrics include the Underwater Colour Image Quality Evaluation Metric (UCIQE)~\cite{55yang2015underwater}, Underwater Image Quality Measure (UIQM)~\cite{56panetta2015human}, Fog Aware Density Evaluator (FADE)~\cite{58choi2015referenceless}, Blind/Referenceless Image Spatial Quality Evaluator (BRISQUE)~\cite{59mittal2011blind}, and Neural Image Assessment (NIMA)~\cite{60talebi2018nima}. Among them, UCIQE and UIQM are specifically designed for underwater image restoration evaluation, while FADE, BRISQUE, and NIMA are commonly used to assess dehazing performance in real-world scenarios. For PSNR, SSIM, UCIQE, UIQM, and NIMA, higher values indicate better performance, whereas lower values are preferred for LPIPS, FADE, and BRISQUE. In the tables, the best and second-best results are highlighted in bold and underlined, respectively.

\subsection{More Experiments}
\label{sec:expset}

\begin{figure*}[t]
    \centering
    \includegraphics[width=\linewidth]{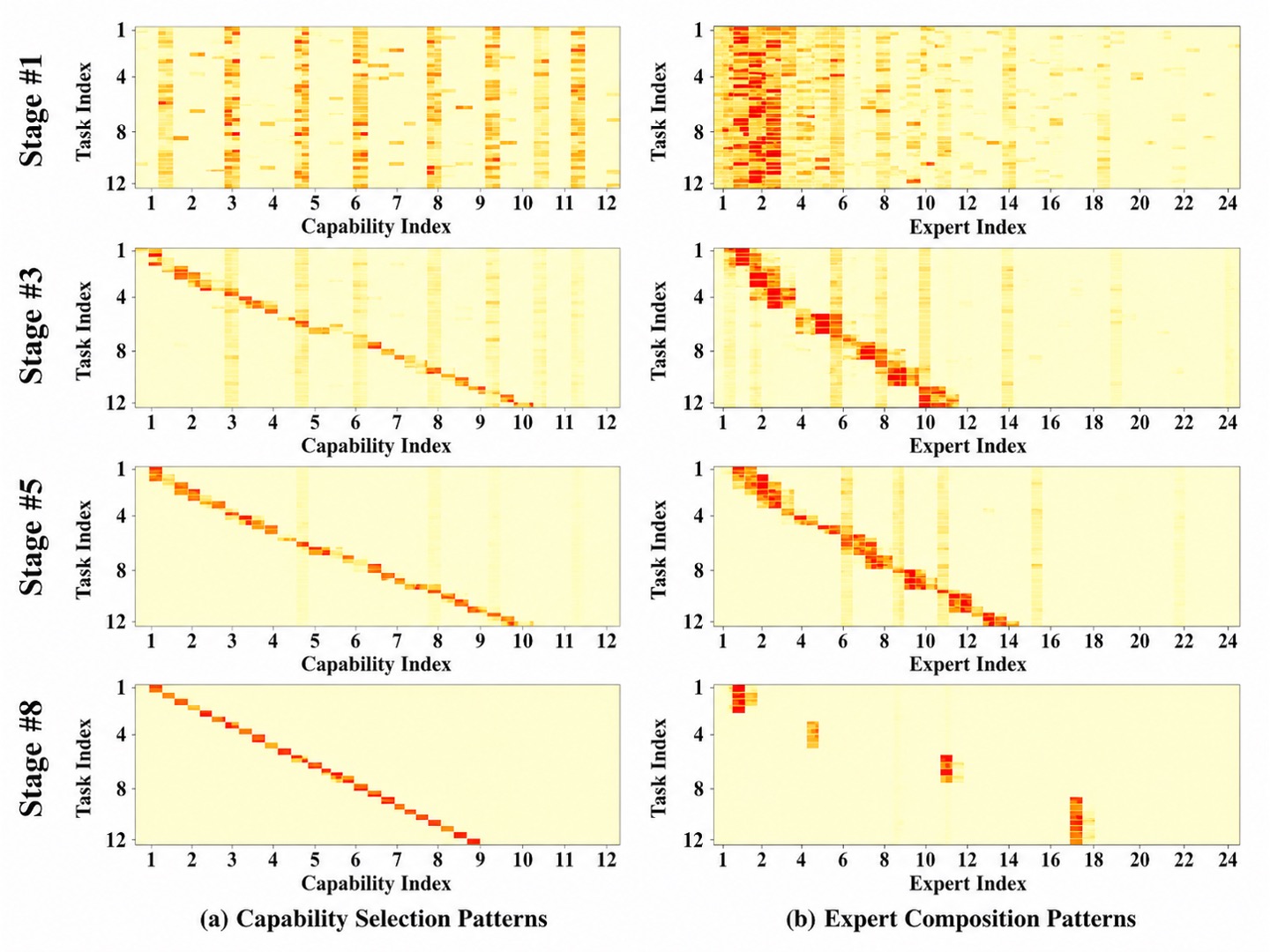}
    \caption{
    Visualization of the stage-wise bi-level routing patterns in RestoreMore.
    The left column shows the capability-selection patterns of the first routing level, while the right column shows the expert-composition patterns of the second routing level.}
    \label{fig:routing_patterns}
\end{figure*}

\subsubsection{Analysis of Bi-level Routing Patterns}
\label{subsec:routing_analysis}

Fig.~\ref{fig:routing_patterns} visualizes the learned routing behaviors of RestoreMore at different feature stages.
The left column shows the capability-selection patterns from the first routing level, and the right column shows the expert-composition patterns from the second routing level.
Several observations can be made.

First, the capability-selection patterns become progressively more structured from shallow to deep stages.
At Stage 1, the routing responses are relatively dispersed, indicating that shallow features tend to retrieve multiple capabilities and emphasize broadly shared low-level restoration cues.
As the stage goes deeper, the activations gradually concentrate around task-aligned regions.
In particular, Stage 3, Stage 5, and Stage 8 exhibit increasingly clear diagonal structures, suggesting that deeper representations are more discriminative and can more accurately identify the capability most relevant to each restoration task.
This behavior is consistent with our motivation that shallow features mainly capture generic statistics and textures, whereas deeper features encode more task-specific restoration semantics.

Second, the expert-composition patterns reveal a transition from shared knowledge reuse to more selective expert specialization.
At Stage 1, a relatively broad set of experts, especially low-index experts, is activated across multiple tasks.
This indicates that early-stage restoration prefers to reuse a set of shared experts that encode transferable low-level priors.
In contrast, the deeper stages show increasingly sparse and localized activations.
At Stage 3 and Stage 5, the activated experts align more clearly with task-dependent regions, while Stage 8 exhibits highly concentrated responses on only a few experts.
This demonstrates that the second routing level indeed performs selective expert composition rather than dense aggregation, enabling the model to preserve shared knowledge while allocating more specialized experts to different tasks.

Third, the overall sparsity of both columns verifies the effectiveness of the proposed bi-level routing mechanism.
Although RestoreMore continually accumulates capability prototypes and experts, each task only activates a small subset of them at each stage.
This observation is consistent with the Top-$M$ capability selection and Top-$K$ expert composition design, which prevents unnecessary interference among unrelated tasks and improves scalability as the supported degradation set grows.

\begin{figure}
    \centering
    \includegraphics[width=.8\linewidth]{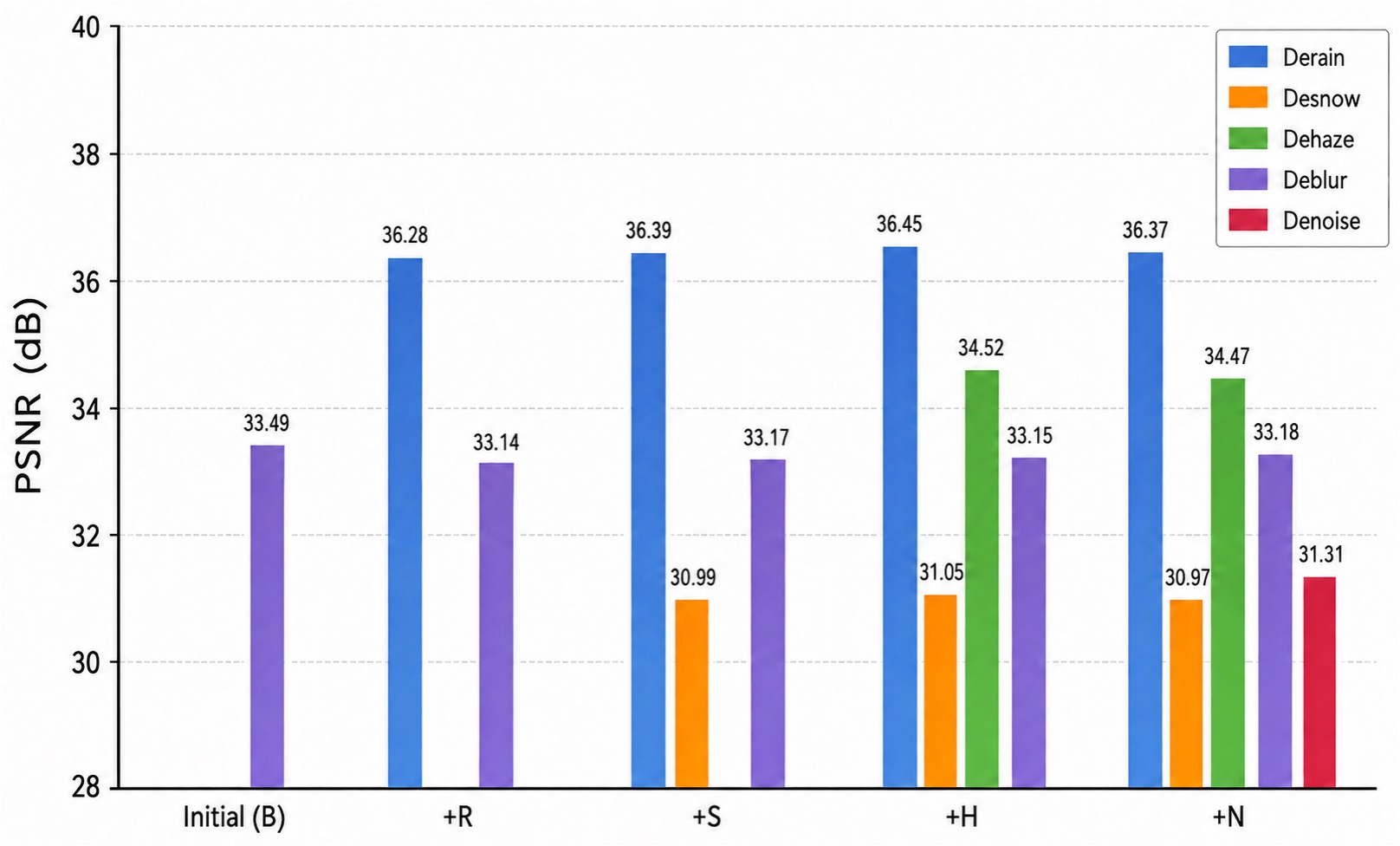}
    \caption{
    Capability evolution of ALGNet under the incremental sequence
    R$\rightarrow$S$\rightarrow$H$\rightarrow$N.
    ALGNet initially supports only deblurring (B), and RestoreMore progressively
    introduces deraining (R), desnowing (S), dehazing (H), and denoising (N).
    Bars report the PSNR of all capabilities available after each incremental stage.
    Previously acquired capabilities remain stable as new restoration tasks are introduced.  }
    \label{fig:algnet_capability_evolution}
\end{figure}

\subsubsection{Capability Evolution during Continual Expansion}
\label{app:capability_evolution}

To further examine how restoration capabilities evolve during continual
learning, Fig.~\ref{fig:algnet_capability_evolution} reports the performance
of ALGNet after each incremental stage under the default
R$\rightarrow$S$\rightarrow$H$\rightarrow$N sequence.
The original ALGNet model supports only deblurring and achieves 33.49 dB.
After deraining is introduced, RestoreMore reaches 36.28 dB on deraining
while retaining 33.14 dB on the original deblurring task, corresponding to
only a 0.35 dB decrease despite acquiring a previously unsupported capability.
As additional tasks are introduced, the previously acquired capabilities
remain highly stable.
After the final denoising stage, the model simultaneously supports all five
restoration tasks, achieving 36.37, 30.97, 34.47, 33.18, and 31.31 dB on
deraining, desnowing, dehazing, deblurring, and denoising, respectively.
Notably, the original deblurring capability decreases by only 0.31 dB from
33.49 to 33.18 dB over four successive capability-expansion stages.
Similarly, once introduced, deraining, desnowing, and dehazing exhibit only
minor variations as subsequent tasks are learned. These results provide a stage-by-stage view of the stability--plasticity behavior of RestoreMore. The model continually acquires previously unsupported restoration capabilities while inducing only limited changes to those learned earlier.

\begin{figure}
    \centering
    \includegraphics[width=\linewidth]{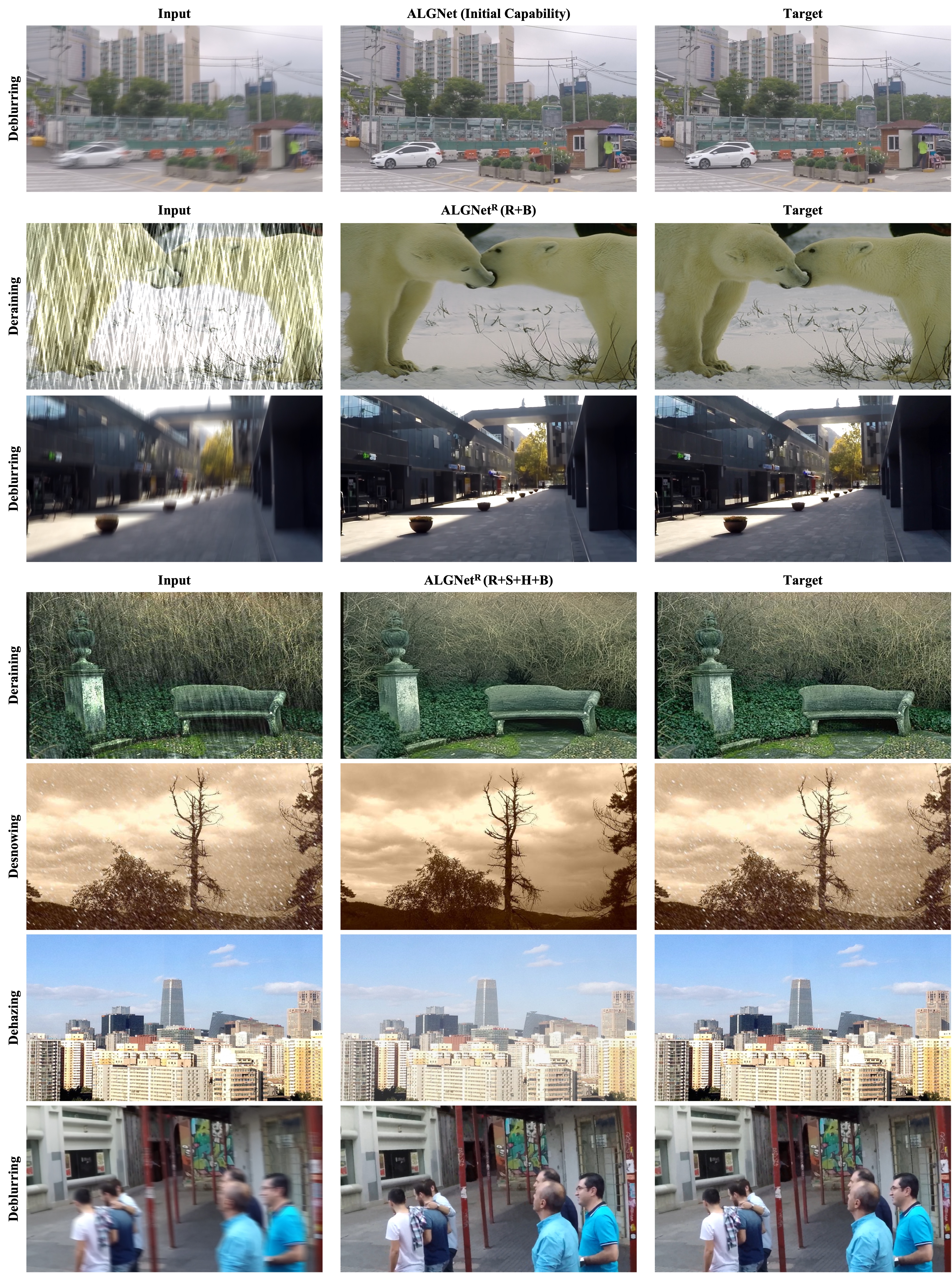}
   \caption{
Qualitative evolution of ALGNet during the early stages of continual capability
expansion.
The original ALGNet initially supports only deblurring, and RestoreMore
progressively introduces additional restoration capabilities.
Representative results from intermediate checkpoints show that newly acquired
tasks are effectively learned while previously available restoration
capabilities remain visually stable.
}
\label{fig:qualitative_evolution_1}
\end{figure}

\begin{figure}
    \centering
    \includegraphics[width=\linewidth]{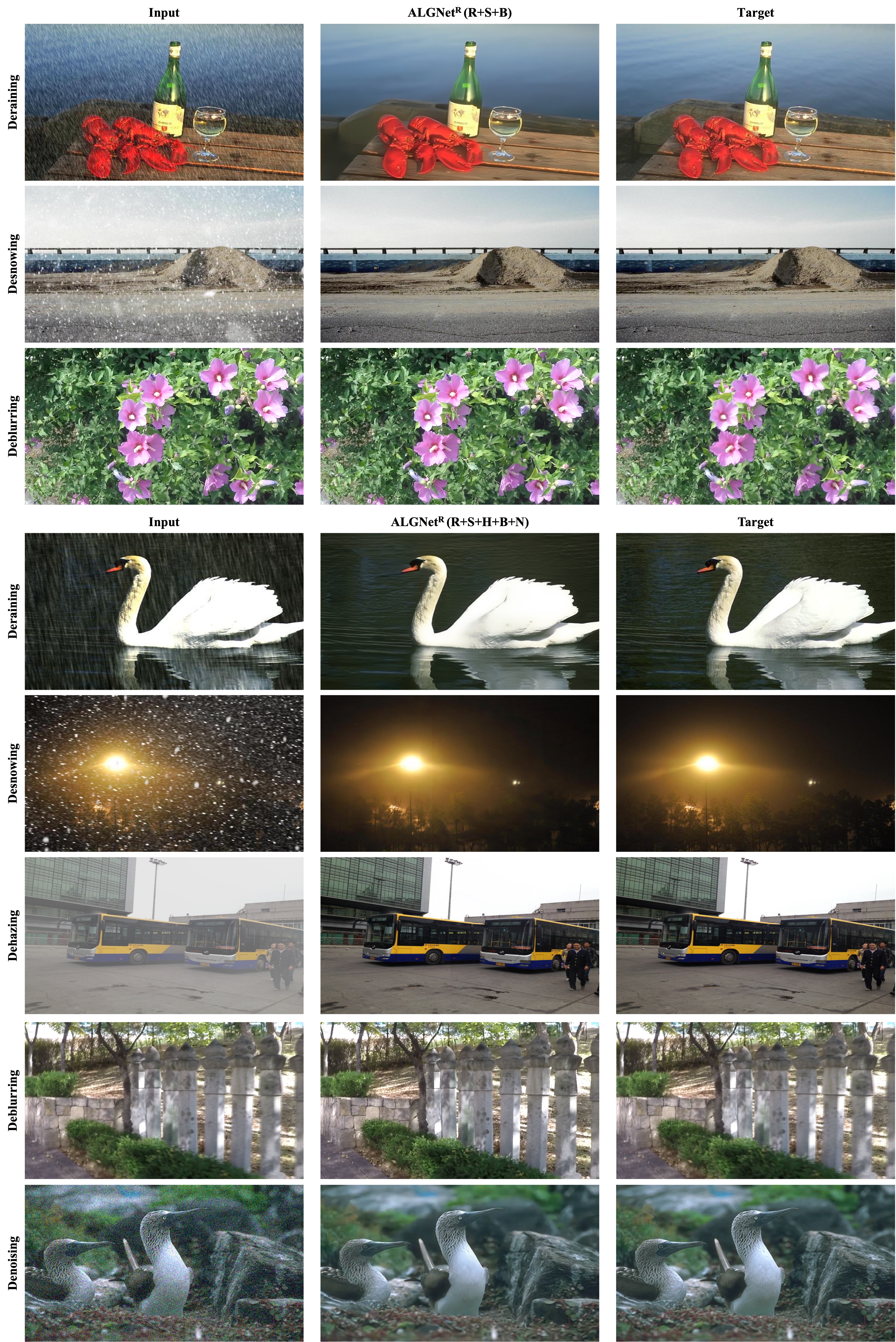}
  \caption{
Qualitative results of ALGNet at later stages of continual capability expansion.
RestoreMore progressively expands the model under the
R$\rightarrow$S$\rightarrow$H$\rightarrow$N sequence, ultimately supporting
deraining, desnowing, dehazing, deblurring, and denoising within a single
expanded model.
}
\label{fig:qualitative_evolution_2}
\end{figure}

Figs.~\ref{fig:qualitative_evolution_1} and~\ref{fig:qualitative_evolution_2}
further visualize how the restoration capability of ALGNet evolves during
continual expansion under the R$\rightarrow$S$\rightarrow$H$\rightarrow$N
sequence.
Starting from the original deblurring model, ALGNet produces a visually sharp
result that closely matches the clean target.
After the first expansion stage, the resulting model jointly supports
deraining and deblurring.
The newly introduced deraining capability effectively suppresses dense rain
streaks while recovering the underlying scene content, whereas the previously
learned deblurring capability remains visually stable, with object boundaries
and local structures still clearly restored.
This provides an intuitive example of acquiring a new capability without
noticeably sacrificing the original restoration pathway.
After desnowing is further introduced, the expanded model successfully removes
snow particles while preserving scene structures and natural appearance.
Meanwhile, the deraining and deblurring results remain consistent with those
obtained at the preceding stage.
A similar behavior is observed after adding dehazing: the model substantially
improves scene visibility and contrast on hazy inputs, while maintaining strong
performance on the previously acquired deraining, desnowing, and deblurring
tasks.
In particular, fine structures such as object contours, vegetation, buildings,
and local textures remain well preserved across the accumulated task set.

The final model, denoted by
ALGNet$^{R}$(R+S+H+B+N), simultaneously supports all five restoration
capabilities.
As shown in Fig.~\ref{fig:qualitative_evolution_2}, the newly acquired
denoising capability effectively suppresses strong noise while retaining
fine object textures.
At the same time, the visual quality of deraining, desnowing, dehazing, and
deblurring remains comparable to the corresponding earlier checkpoints.
For example, rain and snow artifacts are removed without excessive smoothing,
hazy scenes recover clear visibility, and blurred regions retain sharp
structural details.

\end{document}